\documentclass[conference]{IEEEtran}
\IEEEoverridecommandlockouts
\usepackage{amsmath,amssymb,amsfonts}
\usepackage{pgfgantt}
\usepackage{algorithmic}
\usepackage{graphicx}
\usepackage{textcomp}
\usepackage{xcolor}
\def\BibTeX{{\rm B\kern-.05em{\sc i\kern-.025em b}\kern-.08em
    T\kern-.1667em\lower.7ex\hbox{E}\kern-.125emX}}

\usepackage[square,sort,comma,numbers]{natbib}
\usepackage[flushleft]{threeparttable}
\usepackage{hyperref}
\usepackage[symbol]{footmisc}
\usepackage{amsthm}
\usepackage{amsmath}   
{
      \theoremstyle{plain}

      \newtheorem{defn}{Definition}

      \newtheorem{remark}{Remark}
}

\begin{document}

\title{
Distributed Learning And Its Application For Time-Series Prediction}

\author{\IEEEauthorblockN{Nhuong V. Nguyen}
\IEEEauthorblockA{\textit{Dept. Computer Science and Engineering} \\
\textit{University of Connecticut}\\
Connecticut, USA \\
\texttt{nhuong.nguyen@uconn.edu}}
\and
\IEEEauthorblockN{Sybille Legitime}
\IEEEauthorblockA{\textit{Dept. Computer Science and Engineering} \\
\textit{University of Connecticut}\\
Connecticut, USA \\
\texttt{sybille.legitime@uconn.edu}}
}

\maketitle

\begin{abstract}


Extreme events are occurrences whose magnitude and potential cause extensive damage on people, infrastructure, and the environment. Motivated by the extreme nature of the current global health landscape, which is plagued by the coronavirus pandemic, we seek to better understand and model extreme events. Modeling extreme events is common in practice and plays an important role in time-series prediction applications. Our goal is to (i) compare and investigate the effect of some common extreme events modeling methods to explore which method can be practical in reality and (ii) accelerate the deep learning training process, which commonly uses deep recurrent neural network (RNN), by implementing the asynchronous local Stochastic Gradient Descent (SGD) framework among multiple compute nodes. In order to verify our distributed extreme events modeling, we evaluate our proposed framework on a stock data set S\&P500, with a standard recurrent neural network. Our intuition is to explore the (best) extreme events modeling method which could work well under the distributed deep learning setting. Moreover, by using asynchronous distributed learning, we aim to significantly reduce the communication cost among the compute nodes and central server, which is the main bottleneck of almost all distributed learning frameworks.

We implement our proposed work and evaluate its performance on representative data sets, such as S\&P500 stock in $5$-year period. The experimental results validate the correctness of the design principle and show a significant training duration reduction upto $8$x, compared to the baseline single compute node.  Our results also show that our proposed work can achieve the same level of test accuracy, compared to the baseline setting.

\end{abstract}


\section{Introduction and Background}
Events are considered extreme if they have the potential for tremendous impacts on social, ecological, and technical systems~\citep{Qi52}. Such events can vary from a coronal mass ejection during solar storms and volcano eruptions, to extreme rainfall, stock market crashes, and pandemics. As we are currently experiencing an extreme event, namely the pandemic caused by the COVID-19 virus outbreak, our interest grew in understanding the current systems used for time-series and extreme event prediction. Furthermore, we also seek to investigate possible areas of improvement in the system training process.

In this project, we mainly focus on time-series prediction using deep recurrent neural network. Analyzing time-series data has yielded some important applications, such as weather forecasting and stock prediction, which often need to obtain high accuracy under specific safety-critical constraints.

In the scope of this research, we mainly focus on extreme events, which have an interesting property in time-series prediction. In practice, extreme events are usually hard to investigate. This comes from the fact that the number of extreme events is negligible in size, compared to other normal events. Hence, this leads to the problem of \textit{imbalanced data sets}, which still remains an open problem to solve~\citep{ding2019modeling}. Specifically, if we use the standard sampling methods, such as window sliding time, randomized events can lead to \textit{underfitting} issues because an extreme event has a small frequency in appearing as a training sample. However, we can attempt to integrate the events to every single sample, i.e, we duplicate the extreme events to  break the \textit{imbalanced barrier}. By using this trick, our deep learning model can learn the pattern of extreme events occurrences; however, in practice, this trick is shown to suffer from the \textit{overfitting} issue.

Hence, we will do a sensitivity study on how to tackle the problem of \textit{imbalanced data}, especially for time-series data sets. Due to the time constraint, we will conduct our research on some common \textit{imbalanced data} handling tricks, including on the \textit{extreme value theory}, which provides better predictions on future occurrences of extreme events. Note that there is always a trade-off between the efficiency and the effectiveness of \textit{imbalanced data set handling methods}. For instance, if we choose a randomized  time-series sample, we can significantly accelerate the training process, including by using a distributed learning framework while \textit{extreme value theory} is shown to achieve a decent performance, but suffers from the computational cost. Therefore, our questions are: how can we effectively and efficiently parallelize the extreme event modeling method with deep recurrent neural network, and which modeling method we can use?

The above analysis raises the question of the \textit{distributed deep learning framework}. In practice, there is a wide variety of distributed learning frameworks, ranging from distributed multiple cores (process) to distributed multiple compute nodes. Specifically, in distributed multiple cores setting, we use one computer equipped with multiple CPUs or GPUs joining the training process. The information (gradient or the local model) is exchanged through high-speed computer buses while in distributed multiple compute nodes, the training process is run in local machines and the information is exchanged via the Internet or the local LAN network. In this proposal, we mainly use the second architecture, which uses multiple compute nodes. Moreover, from a synchronous standpoint, we can further classify the distributed learning framework into a synchronous and an asynchronous setting, which might contain synchronous points or not. To fully utilize the computation power of the compute nodes, we prefer to use the asynchronous distributed learning framework in this research proposal.

So, we can summarise our contributions in this research proposal as follows:
\begin{enumerate}
    \item We perform a sensitivity study on the \textit{imbalanced data set handling} methods, especially extreme events to investigate which methods we can use in practice.
    
    \item We propose an asynchronous distributed learning for time-series prediction application. Our goal is to show that we can significantly reduce the communication cost while keeping a decent convergence performance, compared to the baseline serial time-series prediction.
\end{enumerate}

\section{Related work}


\subsection{Extreme Event Modeling}

As mentioned in the introduction, many deep neural networks suffer from \textit{overfitting} or \textit{underfitting} in their predictions when trained with \textit{imbalanced data}. D. Ding, M. Zhang et al. coined the term \textit{Extreme Event Problem} ~\citep{ding2019modeling} to refer to this phenomenon.
To establish a formal understanding of the \textit{Extreme Event Problem}, they have introduced an auxiliary indicator sequence~\citep{ding2019modeling} \begin{math} V_{1:T} = [v_1, ..., v_t]:\end{math} \begin{equation}
 v_t = \begin{cases}
      1 & \text{$y_t > \epsilon_1$}\\
      0 & \text{$y_t \in [-\epsilon_2, \epsilon_1] $}\\
      -1 & \text{$y_t < -\epsilon_2$}
    \end{cases}
 \end{equation} 
 with the large constants $\epsilon_1, \epsilon_2 > 0$ as \textit{thresholds}. For time step $t$, it $v_t = 0$, the output $y_t$ is defined as a \textit{normal event}. If $v_t > 0$ the output $y_t$ is defined as a \textit{right extreme event}. In the case of $v_t < 0$ the output $y_t$ is defined as a \textit{left extreme event}. \newline

There is a remarkable correlation between the Extreme Event Problem and the shape of the data distribution. Previous work has established this correlation by noticing that real-world data presents a distribution that always appears to be heavy-tailed. If a random variable $X$ is said to respect a heavy-tailed distribution, then it usually has a non-negligible probability of taking extreme values (larger $\epsilon_1$ and smaller than $-\epsilon_2$).\newline In fact, modeling with light-tailed parametric distributions including Gaussian and Poisson, as opposed to heavy-tailed distributions such as Pareto and log-Cauchy distribution, will bring inevitable losses in the tail of the data. therefore, addressing the Extreme Event Problem involves adequately modeling the distribution of real-world data, especially the tail distribution, to avoid losing valuable extreme event information.

\textbf{Extreme Value Theory (EVT). } Extreme Value Theory takes a step further on studying these heavy-tailed data, making it an important tool for many machine learning and deep learning systems performing extreme event forecasting. Extreme Value Theory focuses on the stochastic behavior of extreme events found in the tails of probability distributions~\citep{showto2016}; more specifically, it studies the distribution of maximum in observed samples. A formal definition of the distribution of the maximum is as follows: suppose T random variables \begin{math} y_1,...,y_T\end{math} are i.i.d sampled from distribution $F_Y$. The distribution of the maximum is,
\begin{equation}
    \lim_{T\to\infty} P\{\max(y_1,...,y_T) \leq y\} = \lim_{T\to\infty} F^T(y) = 0
\end{equation}
To have \begin{math}P\{\max(y_1,...,y_T) \leq y\}\end{math} non-vanishing, a linear transformation is applied on the maximum. This insight derives a fundamental result in EVT, as a theorem that states that the distribution of Y after being linearly transformed is always limited to a few cases~\citep{ding2019modeling}. This theorem represents the Generalized Extreme Value Distribution $G(y)$
\begin{equation}
    G(y) = \begin{cases}
      \exp \left(-(1-\frac{1}{\gamma}y)^\gamma\right) &, \gamma \neq 0,1 - \frac{1}{\gamma}y > 0\\
      \exp\left(-e^{-y}\right) &, \gamma = 0
    \end{cases}
\end{equation}

\textbf{Modeling the Tail in the Distribution. } Work extending the theorem above models the tail distribution of real-world time-series data with the following equation 
\begin{equation}
    1 - F(y) \approx (1 - F(\xi))\left[1-\log G\left(\frac{y-\xi}{f(\xi)}\right)\right], y > \xi
\end{equation} where $y$ represents a random variable sampled from distribution $F$, and $\xi > 0$ is a sufficiently large threshold. Previous researches have shown that the approximation in equation 4 can fit the tail distribution well~\citep{dehaan2006}. Leveraging the EVT to effectively model tailed distributions will alleviate the Extreme Event Problem that deep neural networks often encounter.

\textbf{Extreme Value Loss.} Work to use Extreme Value Theory as inspiration to impose approximated tailed distribution on observations provided a classification called Extreme Value Loss (EVL)~\citep{ding2019modeling}. To define the Extreme Value Loss function, rather than directly modeling the output, we refer to the extreme event indication defined in equation 1. Using the tail modeling approximation in equation 4, we can write the approximation for right extreme event $y_t$ as
\begin{equation}
    1 - F(y) \approx (1 - P(v_t = 1))\log G\left(\frac{y_t-\epsilon_1}{f(\epsilon_1)}\right)
\end{equation}, where positive function $f$ is the scale function. If we consider a binary classification task for determining right extreme events, we can set our predicted indicator as $u_t$ which can be seen as a hard approximation for \begin{math}(y_t - \epsilon_1)/f(\epsilon_1)\end{math}. The approximation represents weights that we add on each term in binary cross entropy,
\begin{equation}
    \begin{split}
        EVL(u_t) = & -\beta_0\left[1-\frac{u_t}{\gamma}\right]^\gamma v_t\log(u_t) \\
        - &\beta_1\left[1-\frac{1 - u_t}{\gamma}\right]^\gamma (1-v_t)\log(1 - u_t)
    \end{split}
\end{equation}, where $\beta_0 = P(v_t = 0)$ which is the proportion of normal events in the dataset and  $P(v_t = 1)$ is the proportion of right extreme events in the dataset.$\gamma$ represents the \textit{extreme value index} in the approximation and is a hyper-parameter. This classification loss function is the Extreme Value Loss.The term $\beta_0$ will increase the penalty when the model recognizes the event $y_t$ as a normal event, and the term $[1-u_t/\gamma]^\gamma$ will also increase the penalty when the model recognizes the extreme event with little confidence, effectively finding the proper weights by adding approximation on the tail distribution of observations through Extreme Value Theory.

Previous work has introduced a deep neural network framework which combines a memory network module which uses GRU for memorizing the characteristics of extreme events, and the Extreme Value Loss, for modeling tail distribution. While we are not considering integrating the memory network module in this study, We find there to be promise in incorporating the Extreme Value Loss to our distributed deep learning framework to achieve better prediction performance for extreme events in time-series data.

\textbf{Another Paradigm for Extreme Event Prediction. } Genetic Programming, a type of evolutionary algorithm, is used to discover solutions to problems which are hard to solve by the best human efforts. Inspired by biological evolution and its fundamental mechanisms, GP software systems implement an algorithm that uses random mutation, crossover, a fitness function, and multiple generations of evolution to resolve a user-defined task. In the context of time-series, genetic algorithms have been used for stock price prediction~\citep{sachdev2015}, and for extreme event predictions in particular, genetic programming models have been implemented to downscale extreme rainfall events, showing reasonable accuracy~\citep{hadipour2013gp}. Although we will not, in the context of this project, delve into investigating in detail the fitness of genetic algorithms, it is noteworthy to mention other types of algorithms are designed to handle the \textit{extreme event problem}.

\subsection{Deep Learning time-series Prediction}


\textbf{Traditional time-series method.} Traditional methods such as autoregressive moving average (ARMA)~\citep{moran1951hypothesis} and nonlinear autoregressive exogenous (NARX)~\citep{lin1996learning} use statistical models with few parameters to exploit patterns in time-series data.

\textbf{Deep recurrent neural network.} Recently, with the celebrated success of Deep Neural Networks (DNN) in many fields such as image classification~\citep{krizhevsky2012imagenet}, a number of DNN-based techniques have been subsequently developed for time-series prediction tasks, achieving noticeable improvements over traditional methods~\citep{dasgupta2017nonlinear}. As a basic component of these models, the Recurrent Neural Network (RNN) module serves as an indispensable factor for these note-worthy improvements~\citep{yan2013substructure}. Compared with traditional methods, one of the major advantages of RNN structure is that it enables deep non-linear modeling of temporal patterns. In recent literature, some of its variants show even better empirical performance, such as the well-known Long-Short Term Memory (LSTM) and Gated Recurrent Unit (GRU)~\citep{hochreiter1997long,chung2014empirical}, while the latter appears to be more efficient on smaller and simpler dataset. 

\textbf{Imbalanced data sets.} Most previously studied DNN are observed to have trouble in dealing with data imbalance~\citep{wang2019neural,ding2019modeling}. Illustratively, let us consider a binary classification task with its training set including 99\% positive samples and only 1\% negative samples, which is said to contain data imbalance. Moreover, imbalance in data will potentially bring any classifier into either of two unexpected situations: a. the model hardly learns any pattern and simply chooses to recognize all samples as positive. b. the model memorizes the training set perfectly while it generalizes poorly to test set.

\subsection{Distributed Deep Learning}

\textbf{Synchronous SGD.} Local SGD~\citep{stich2018local} or Federated Learning (FL) \citep{mcmahan,bonawitz2019towards} is a distributed machine learning approach which enables training on a large corpus of decentralized data located on devices like mobile phones or IoT devices. While local SGD does not care about the data privacy, federated learning assumes that there is no data shared between any client to prevent data leakage. Original FL requires synchrony between the server and clients (compute nodes).  It requires that each client send a full model back to the server in each round and that each client wait for the next computation round. For large and complicated models, this becomes a main bottleneck  due to the asymmetric property of internet connection and the different computation power of devices \citep{yang,konevcny2016federated}.

\textbf{Asynchronous SGD.} Asynchronous training~\citep{lian2015asynchronous,zheng2017asynchronous} is widely used in traditional distributed SGD. Hogwild!, one of the most famous asynchronous SGD algorithms, was introduced in ~\citep{Hogwild} and various variants with a fixed and diminishing step size sequences were introduced in \citep{DeSaZhangOlukotunEtAl2015,Leblond2018,nguyen2018sgd}.
%
Typically, asynchronous training converges faster than synchronous training in real time due to parallelism. This is because in a synchronized solution compute nodes have to wait for the slower ones to communicate their updates after which a new global model can be downloaded by everyone. This causes high idle waiting times at compute nodes.
Asynchronous training allows compute nodes to continue executing SGD recursions based on stale global models. For non-convex problems, synchronous training \citep{Ghadimi2013Minibatch} and  asynchronous training with bounded staleness \citep{lian2015asynchronous} - or in our terminology, bounded delay - achieves the same convergence rate of $O(1/\sqrt{nK})$, where $n$ is the number of compute nodes and $K$ is the total number of gradient computations.
Another work from~\citep{van2020hogwild} introduces the asynchronous SGD with increasing sample sizes to significantly reduce the communication cost. This work analyses and demonstrates the promise of this new technique with the convergence rate guarantee for both i.i.d and heterogeneous data sets.

\textbf{Event-triggered SGD.} Event-triggered SGD is a special case of asynchronous SGD, where each client independently computes the SGD recursions and broadcast their SGD computation only to their neighbors. The event-triggered threshold plays an important role in this algorithm since it decides the number of communication cost among clients and the convergence performance. While the work from~\citep{george2019distributed} measures the difference between the current local model and the new one as a threshold to broadcast their new models, other works~\citep{hsieh2017gaia,luping} use the relevant gradient metric from the clients to decide the period when this SGD computation can be sent to other neighbors.

\begin{figure}[ht!]
\includegraphics[width=1.0\columnwidth]{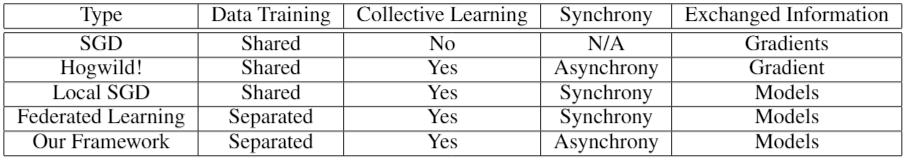}
\caption{Summary information of SGD-based methods for distributed learning setting.}
\label{fig:summary}
\end{figure}

\begin{figure}[ht!]
\centering
\includegraphics[width=0.6\columnwidth]{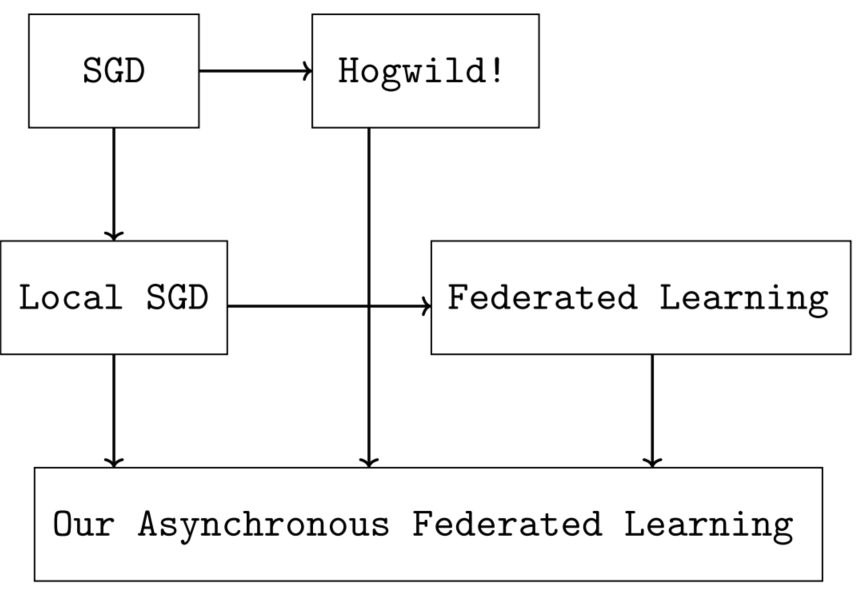}
\caption{Related work and the relationship with our work.}
\label{fig:related_work}
\end{figure}

We summarise the setting and information exchanged of some related distributed learning as in Figure~\ref{fig:summary} and Figure~\ref{fig:related_work}. Note that the information exchanged here can be gradients or models. In our work, we also verify which information exchanged can be used with time-series data and deep learning models.

\section{Proposed directions of technical components}

\subsection{System design}
\begin{figure}[ht!]
\includegraphics[width=0.99\columnwidth]{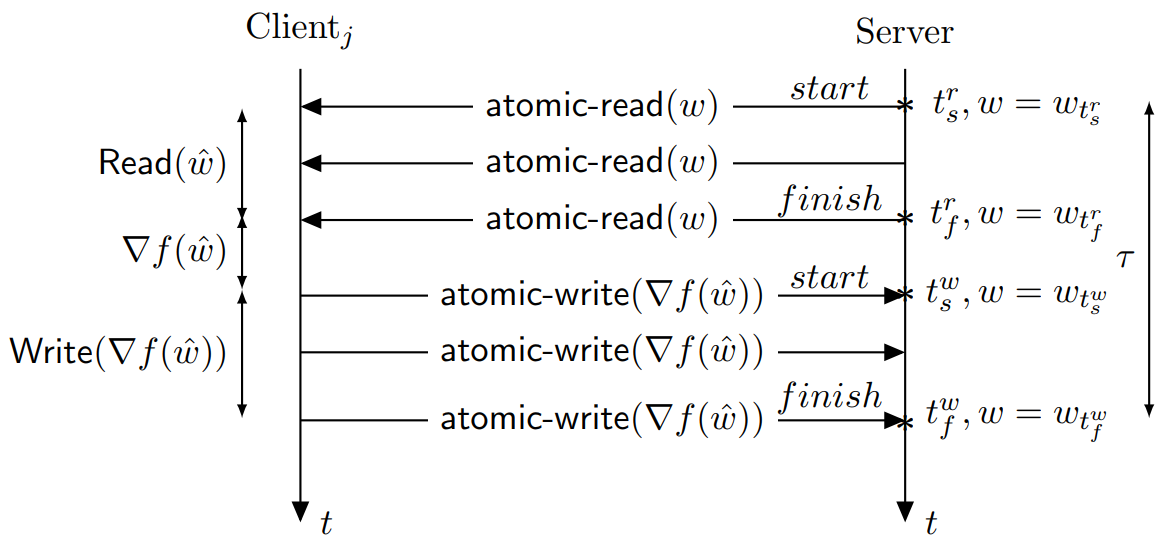}
\caption{Hogwild! algorithm in timeline with delay $\tau$}
\label{fig:hogwild_timeline}
\end{figure}

Our approach is based on the Hogwild!~~\citep{robbins1951stochastic,Hogwild, van2020hogwild}
recursion 
\begin{equation}
 w_{t+1} = w_t - \eta_t  \nabla f(\hat{w}_t;\xi_t),\label{eqwM2a}
 \end{equation}
 where $\hat{w}_t$ represents the vector used in computing the gradient $\nabla f(\hat{w}_t;\xi_t)$ and whose vector entries have been read (one by one)  from  an aggregate of a mix of  previous updates that led to $w_{j}$, $j\leq t$.
 In a single-thread setting where updates are done in a fully consistent way, i.e. $\hat{w}_t=w_t$, yields SGD with diminishing step sizes $\{\eta_t\}$. The detailed illustration of Hogwild! algorithm can be found at Figure~\ref{fig:hogwild_timeline}.
 
Recursion (\ref{eqwM2a}) models asynchronous SGD.
We define the amount of asynchronous behavior by function $\tau(t)$:

\begin{defn}\label{def_delay_tau}
We say that the sequence $\{w_t\}$  is {\em consistent with a delay function $\tau$}  
if, for all $t$, vector $\hat{w}_t$ includes the aggregate 
of the updates up to and including those made during the $(t-\tau(t))$-th iteration\footnote{ (\ref{eqwM2a}) defines the $(t+1)$-th iteration, where $\eta_t\nabla f(\hat{w}_t;\xi_t)$ represents the $(t+1)$-th update.}, i.e., $\hat{w}_t = w_0 - \sum_{j\in {\cal U}} \eta_j \nabla f(\hat{w}_j;\xi_j)$ for some ${\cal U}$ with $\{0,1,\ldots, t-\tau(t)-1\}\subseteq {\cal U}$.
\end{defn}

Our main insight is that the asynchronous SGD framework based on Hogwild! can resist much larger delays than the natural delays caused by the network communication infrastructure, in fact, it turns out that $\tau(t)$ can scale as much as $\approx \sqrt{t/\ln t}$ for strongly convex problems \citep{nguyen2018sgd,nguyen2018new}. Here, $t$ is the current iteration and our clients use the diminishing step size scheme. This means that recurrence (\ref{eqwM2a}) can be used/exploited in an asynchronous SGD implementation over distributed local data sets where much more {\em asynchronous behavior is introduced  by design}. Moreover, the work from~\citep{van2020hogwild} guarantees that rather than having each round  execute the same/constant number of SGD recursions, we can {\em increase the number of SGD iterations performed locally at each compute node  from round to round}. This {\em reduces the amount of network communication} compared to the straightforward usage of recurrence (\ref{eqwM2a}) where each compute node performs a fixed number of SGD iterations within each round.

\begin{remark}{Linearly increasing sample sequences:} \label{rmk:linear_increasing}
From theory and simulation~\citep{van2020hogwild}, in order to bootstrap convergence, it is best  to start the training process with larger step sizes and start with  rounds of small size (measured in the number of local SGD recursions). After that, these rounds should start increasing in  size and should start using smaller and smaller step sizes for best performance (in terms of minimizing communication cost).
In practice, for non-convex problems, we can choose the sample size sequence $s_i = \mathcal{O}(i)$, where $i$ is the current index of the communication round and the corresponding diminishing step size sequence follows $\bar{\eta_i} \sim \mathcal{O}(1 / \sqrt{t})$, where $t$ is the number of iterations until communication round $i$-th. We can observe that for a fixed number of gradient computations $K$, the number $T$ of communication rounds satisfies $K=\sum_{j=0}^T s_j$.
This makes $T$ proportional to $\sqrt{K}$, rather than proportional to $K$ for a constant sample size sequence. 
\end{remark}

\subsection{Overall system design}
\begin{figure}[ht!]
\includegraphics[width=0.99\columnwidth]{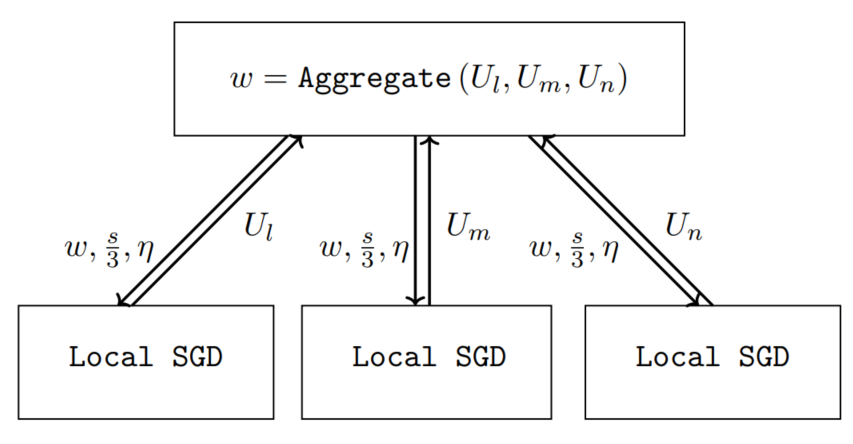}
\caption{Asynchronous distributed learning framework}
\label{fig:async_framework}
\end{figure}

Our overall system design is mainly based on the distributed learning framework, shown as in Figure~\ref{fig:async_framework} and the \textit{linear increasing sample SGD recursions} from the work~\cite{van2020hogwild} and the delay $\tau$ from the Hogwild! algorithm~\cite{Hogwild}.

As can be seen from Figure~\ref{fig:async_framework}\footnote{The server and $n$ clients can work in asynchronous fashion with i.i.d or heterogeneous datasets. At the client side, after running each $s_i/n$-iteration SGD with stepsize $\eta_i$. The audience is referred to the paper~\cite{van2020hogwild} for a precise description of the algorithm.}, we will use multiple compute nodes joining the training process. Here, each compute node can have its own local data set (stock data set in our case) or can share the same data sets. Moreover, each node will run their local SGD independently with other compute nodes. In other words, each node will run one RNN model and exchange the information (gradient or local model) with the central server independent with other compute nodes. Moreover, the compute nodes need to follow the delay $\tau$ constraints in Definition~\ref{def_delay_tau}.


Other settings, such as the RNN model, the extreme event threshold, the attention layer for RNN model are mainly referenced from the work~\citep{ding2019modeling}. We will specify any modification or changes, compared to the baseline method~\citep{ding2019modeling} in our report if possible.

Also, we can see two main parts from Figure~\ref{fig:async_framework}: the compute node (client side) and the central server. Our asynchronous setting is mainly conducted on the compute node's side. In order to avoid repeating, the pseudo code for the central server and the compute node can be found in Algorithm~4 and Algorithm~5 from the work~\cite{van2020hogwild}, respectively.




\section{Experiments}

\subsection{Environmental setting}
For simulating the asynchronous distributed learning with extreme event modeling, we use multiple threads where each thread represents one clients joining the training process. The experiments are mainly conducted on Linux-64bit OS, with $16$ cpu processor, $32$Gb RAM and $2$ NVIDIA GPUs.


\subsection{Data set}
The data this example will be using is the S\&P500 file in the data folder\footnote{https://github.com/jaungiers/LSTM-Neural-Network-for-Time-Series-Prediction/tree/master/data}. This file contains the Open, High, Low, Close prices as well as the daily Volume of the S\&P500 Equity Index from January 2012 to September 2017. We split the subset of data set from 2012 to 2014 as training data sets and the subset of data set from 2015 to 2016 as testing data sets. The chosen stocks are GOOGL, FB, AAPL, AMZN, IBM, NFLX, EBAY. In this report, we only show the experimental results for AAPL and AMZN stock.

\subsection{Experimental tasks}

From the above analysis, our goal is try to explore which is the best way to handle \textbf{imbalanced data sets}, and extreme events is our user case. And then we focus on distributing the training process among multiple compute nodes, in order to reduce not only the training duration but also the communication cost. Hence, our experiments can be as follows:
\begin{enumerate}
    \item Sensitivity study on extreme events modeling.
    \item Implementing and adapting the standard RNN model with extreme events modeling.
    \item Comparing the effective of the extreme events modeling methods, in terms of the prediction accuracy.
    \item Distributing the RNN model with multiple compute nodes. Our goal is to show the significantly commnucation cost reduction, compared to the baseline local SGD method~\citep{stich2018local}.
    
\end{enumerate}

\subsection{Experimental results}
\noindent \textbf{Simulation environment.} For simulating our proposed distributed learning framework, we mainly run our experiments on virtual Linux-64bit machine, with $2$ cpu processors, $2$Gb RAM and $1$ NVIDIA GPU. We use Pytorch library to implement our proposed framework\footnote{The simulation source code can be found at \url{https://drive.google.com/file/d/1eaWL8rMP5Un3d8J88JWBuQG7qmvZAvvX/view?usp=sharing}}. The baseline method is traditional LSTM on single machine. We will compare our work with the baseline in terms of \textit{prediction accuracy} and \textit{training duration}.

\vspace{0.025in}
\noindent \textbf{Objective function.} Our experiments mainly focus on non-convex problems (deep neural networks). For simplicity, we choose a simple LSTM network for stock prediction application\footnote{The deep network model we used here is Input Layer -- 2 LSTM Layers -- 3 Fully Connected Layers for prediction.}. The basic parameter setting is summarized in Table~\ref{tbl:basic_param_setting}.

\begin{table}[ht!]
\caption{Summary of the parameter setting}
\vspace{-0.1in}
\vskip 0.1in
\label{tbl:basic_param_setting}
\begin{center}
\scalebox{1.09}{
\begin{threeparttable}
\begin{tabular}{|c|c|}
\hline
Parameter & Value \\ \hline \hline
Initial stepsize $\bar{\eta}_0$  & $0.01$
\\ \hline
Stepsize  scheme & $\textstyle \bar{\eta}_i = \textstyle \frac{\bar{\eta}_0}{1 + \beta \cdot \sqrt{t}}, \beta = 0.01$
\\ \hline
\# of data points $N_c$ &    $\approx4000$ 
\\ \hline
\# of iterations $K$ &    $\{288375\}$
\\ \hline
Linear sample size $s_i$ &    $s_i = a \cdot i^p + b$ ($a=10, p=1, b=0$)
\\ \hline
Data set & \texttt{S\&P500} stock
\\ \hline
 \# of nodes $n$ & $\{1, 2, 5, 10 \}$
\\ \hline
Sliding window & $20$
\\ \hline

Regularization $\lambda$ & $1/N_c$
\\ \hline

\end{tabular}
  \end{threeparttable}
  
 }
\end{center}
\end{table}

\begin{figure}[ht!]
\includegraphics[width=0.99\columnwidth]{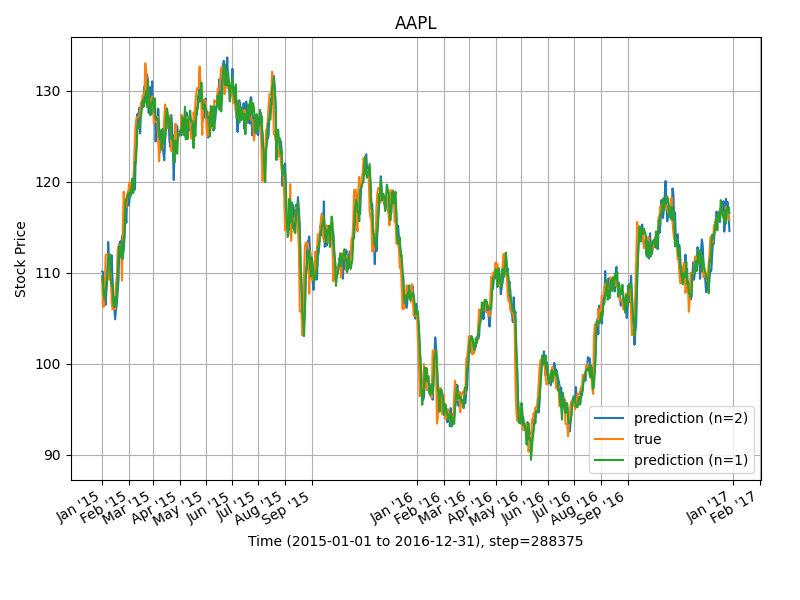}
\caption{Stock prediction of our work for $2$ compute nodes (Apple stock).}
\label{fig:appl_2node}
\end{figure}

\begin{figure}[ht!]
\includegraphics[width=0.99\columnwidth]{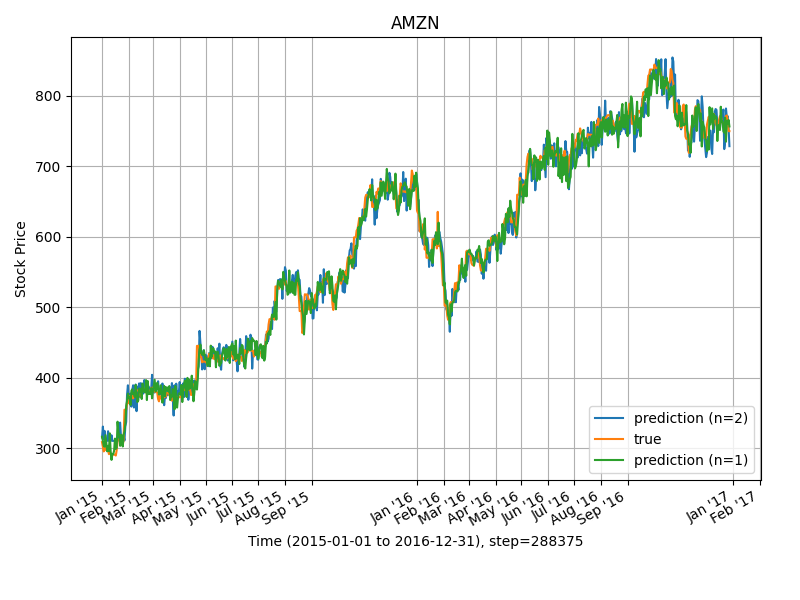}
\caption{Stock prediction of our work for $2$ compute nodes (Amazon stock).}
\label{fig:amazon_2node}
\end{figure}

\begin{figure}[ht!]
\includegraphics[width=0.99\columnwidth]{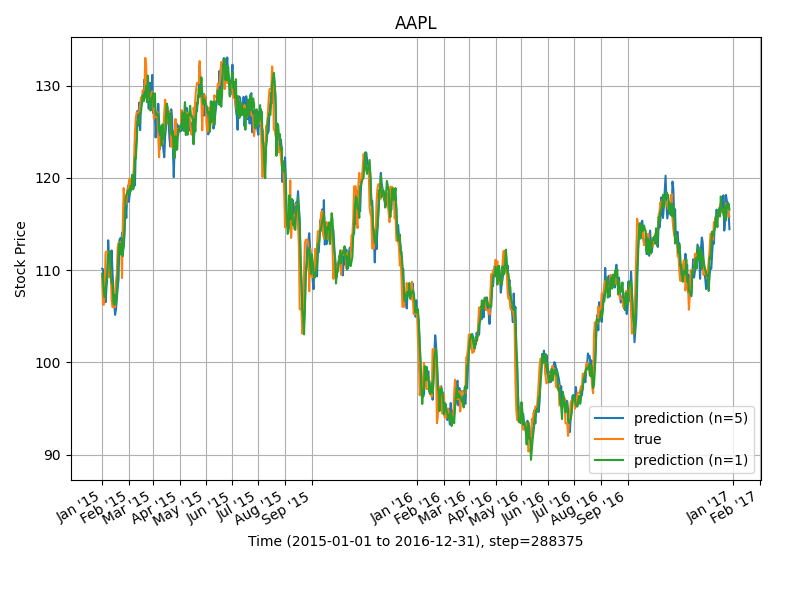}
\caption{Stock prediction of our work for $5$ compute nodes (Apple stock).}
\label{fig:appl_5node}
\end{figure}

\begin{figure}[ht!]
\includegraphics[width=0.99\columnwidth]{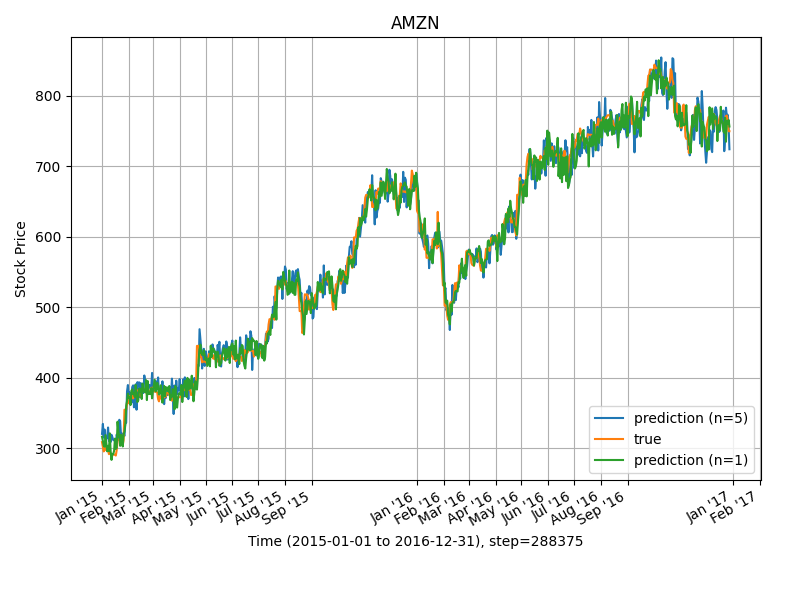}
\caption{Stock prediction of our work for $5$ compute nodes (Amazon stock).}
\label{fig:amazon_5node}
\end{figure}

\begin{figure}[ht!]
\includegraphics[width=0.99\columnwidth]{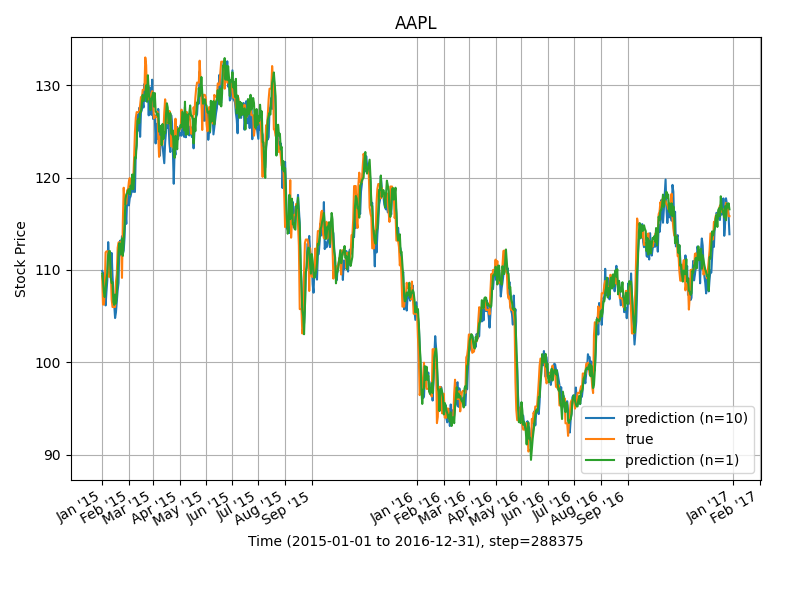}
\caption{Stock prediction of our work for $10$ compute nodes (Apple stock).}
\label{fig:appl_10node}
\end{figure}

\begin{figure}[ht!]
\includegraphics[width=0.99\columnwidth]{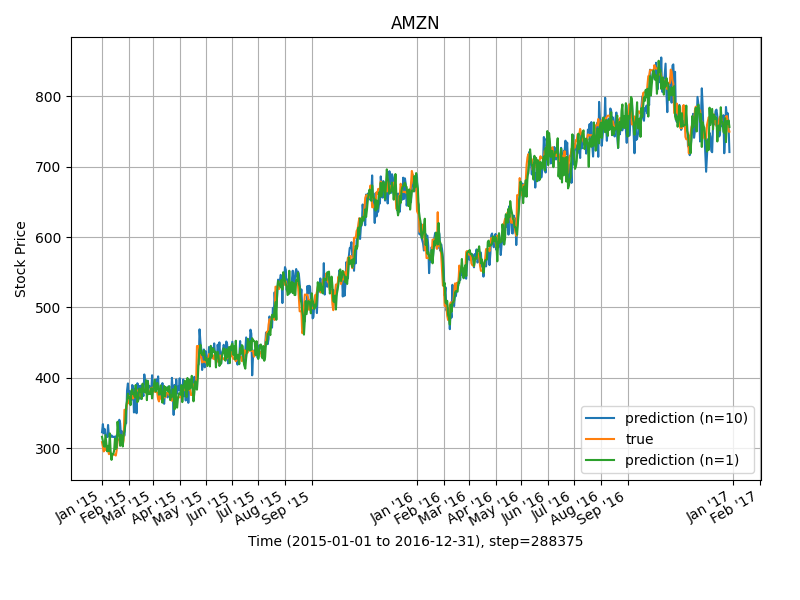}
\caption{Stock prediction of our work for $10$ compute nodes (Amazon stock).}
\label{fig:amazon_10node}
\end{figure}

The prediction performance of our work are shown from Figure~\ref{fig:appl_2node} to Figure~\ref{fig:amazon_10node}. As can be seen here, we can obtain the same level of prediction accuracy, compared to the baseline with single compute nodes. These figures again confirm the effectiveness of our proposed work\footnote{The experimental results for other stocks can be found at \url{https://drive.google.com/drive/folders/1j7kssKkaDjt66fHugOEGu_ZmdxjnSz1a?usp=sharing}}.

\begin{table}[ht!]
\caption{Summary of speedup ratio for different number of compute nodes, compared to the baseline – single compute node.}
\label{tbl:speedup_appl}
\centering
\scalebox{1.25}{
\begin{tabular}{|c|c|c|}
\hline
No. of compute node & Speedup   & Notes \\ \hline
\hline
2                   & $\sim$1.5 &       \\ \hline
5                   & $\sim$4.2 &       \\ \hline
10                  & $\sim$8.3 &       \\ \hline
\end{tabular}
}
\end{table}

Turing to the speedup ratio, from Table~\ref{tbl:speedup_appl}, we can see that the speedup ratio increases when we increase the number of compute nodes. However, the speedup ratio does not exactly scale up to the number of compute nodes. This can explained by the reason that the central server needs to execute the aggregation operation (of local models) to issue new global model for the next computation round. The results confirm that there is a speedup saturation in the distributed learning frameworks. In practice, we need to combine with other methods to increase the bound of speedup saturation.
\section{Discussion}

\subsection{Advantages}
From this work, we gain some achievements as follows:
\begin{enumerate}
    \item {We successfully implement the distributed learning for time-series prediction.}
    
    \item {Our proposed work can significantly reduce the communication cost by using linearly increasing sample sequences and verify its effectiveness by the experimental results.}
    
    \item {We also do a sensitivity study on the extreme events. This work can be a reference for anyone who start to work on the time-series prediction with attention of extreme event.}
\end{enumerate}

\subsection{Disadvantages}
However, due to time limit and experimental environment, we have some limitation as below:
\begin{enumerate}
    \item {We are trying to integrate the extreme event loss to the deep learning model. Because this integration is in its earliest stages, we do not have a chance to measure and report the effectiveness of extreme event loss to the prediction accuracy. }
    
    \item {Our experiments are the simulation artifact. Moreover, the perspective of asynchronous setting is not considered yet. We will extend this kind of experiment when possible.}
    
    \item {The data set S\&P500 is quite simple, we also need to run the experiments for other time-series data sets to verify the performance of our work.}
\end{enumerate}

\section{Conclusion and Future work}
In conclusion, we can ($i$) speedup the training process by using multiple compute nodes. Moreover, the communication cost can be significantly reduced by using linearly increasing sample sequences. We also ($ii$) do a sensitivity study on extreme events and its application for time-series prediction.
We also note that ($iii$) the information exchange in our work is the model, not the gradient information. It seems that the distributed learning framework with gradient exchange does not work well for time-series prediction\footnote{This is the observation from our experiments. It seems that model averaging is better than gradient averaging for time-series prediction deep learning models. This observation can be explained by the gradient vanishing or gradient explosion for RNN models, which makes the gradient averaging less practical in this case. In practice, we can use gradient clipping or better activation functions to alleviate this issue.}. In nearest future, we plan to extend the experiments for asynchronous settings as well as integrate our proposed framework with the extreme event loss function.






\small
\bibliographystyle{IEEEtran}
\bibliography{sampleBibFile}

\begin{thebibliography}{10}
\providecommand{\url}[1]{#1}
\csname url@samestyle\endcsname
\providecommand{\newblock}{\relax}
\providecommand{\bibinfo}[2]{#2}
\providecommand{\BIBentrySTDinterwordspacing}{\spaceskip=0pt\relax}
\providecommand{\BIBentryALTinterwordstretchfactor}{4}
\providecommand{\BIBentryALTinterwordspacing}{\spaceskip=\fontdimen2\font plus
\BIBentryALTinterwordstretchfactor\fontdimen3\font minus
  \fontdimen4\font\relax}
\providecommand{\BIBforeignlanguage}[2]{{%
\expandafter\ifx\csname l@#1\endcsname\relax
\typeout{** WARNING: IEEEtran.bst: No hyphenation pattern has been}%
\typeout{** loaded for the language `#1'. Using the pattern for}%
\typeout{** the default language instead.}%
\else
\language=\csname l@#1\endcsname
\fi
#2}}
\providecommand{\BIBdecl}{\relax}
\BIBdecl

\bibitem{Qi52}
\BIBentryALTinterwordspacing
D.~Qi and A.~J. Majda, ``Using machine learning to predict extreme events in
  complex systems,'' \emph{Proceedings of the National Academy of Sciences},
  vol. 117, no.~1, pp. 52--59, 2020. [Online]. Available:
  \url{https://www.pnas.org/content/117/1/52}
\BIBentrySTDinterwordspacing

\bibitem{ding2019modeling}
D.~Ding, M.~Zhang, X.~Pan, M.~Yang, and X.~He, ``Modeling extreme events in
  time series prediction,'' in \emph{Proceedings of the 25th ACM SIGKDD
  International Conference on Knowledge Discovery \& Data Mining}, 2019, pp.
  1114--1122.

\bibitem{showto2016}
\BIBentryALTinterwordspacing
S.~Glen, ``Extreme value distribution and the extreme value theory,'' 2016,
  last accessed 05/07/2021. [Online]. Available:
  \url{https://www.statisticshowto.com/extreme-value-distribution/}
\BIBentrySTDinterwordspacing

\bibitem{dehaan2006}
L.~de~Haan and A.~Ferreira, \emph{Extreme Value Theory: An Introduction}.\hskip
  1em plus 0.5em minus 0.4em\relax Springer-Verlag New York, 2006, vol.~1,
  no.~1.

\bibitem{sachdev2015}
A.~{Sachdev} and V.~{Sharma}, ``Stock forecasting model based on combined fuzzy
  time series and genetic algorithm,'' in \emph{2015 International Conference
  on Computational Intelligence and Communication Networks (CICN)}, 2015, pp.
  1303--1307.

\bibitem{hadipour2013gp}
S.~{Hadipour}, S.~{Shahid}, S.~B. {Harun}, and X.~{Wang}, ``Genetic programming
  for downscaling extreme rainfall events,'' in \emph{2013 1st International
  Conference on Artificial Intelligence, Modelling and Simulation}, 2013, pp.
  331--334.

\bibitem{moran1951hypothesis}
P.~Moran, ``Hypothesis testing in time series analysis,'' 1951.

\bibitem{lin1996learning}
T.~Lin, B.~G. Horne, P.~Tino, and C.~L. Giles, ``Learning long-term
  dependencies in narx recurrent neural networks,'' \emph{IEEE Transactions on
  Neural Networks}, vol.~7, no.~6, pp. 1329--1338, 1996.

\bibitem{krizhevsky2012imagenet}
A.~Krizhevsky, I.~Sutskever, and G.~E. Hinton, ``Imagenet classification with
  deep convolutional neural networks,'' \emph{Advances in neural information
  processing systems}, vol.~25, pp. 1097--1105, 2012.

\bibitem{dasgupta2017nonlinear}
S.~Dasgupta and T.~Osogami, ``Nonlinear dynamic boltzmann machines for
  time-series prediction,'' in \emph{Proceedings of the AAAI Conference on
  Artificial Intelligence}, vol.~31, no.~1, 2017.

\bibitem{yan2013substructure}
L.~Yan, A.~Elgamal, and G.~W. Cottrell, ``Substructure vibration narx neural
  network approach for statistical damage inference,'' \emph{Journal of
  Engineering Mechanics}, vol. 139, no.~6, pp. 737--747, 2013.

\bibitem{hochreiter1997long}
S.~Hochreiter and J.~Schmidhuber, ``Long short-term memory,'' \emph{Neural
  computation}, vol.~9, no.~8, pp. 1735--1780, 1997.

\bibitem{chung2014empirical}
J.~Chung, C.~Gulcehre, K.~Cho, and Y.~Bengio, ``Empirical evaluation of gated
  recurrent neural networks on sequence modeling,'' \emph{arXiv preprint
  arXiv:1412.3555}, 2014.

\bibitem{wang2019neural}
X.~Wang, X.~He, M.~Wang, F.~Feng, and T.-S. Chua, ``Neural graph collaborative
  filtering,'' in \emph{Proceedings of the 42nd international ACM SIGIR
  conference on Research and development in Information Retrieval}, 2019, pp.
  165--174.

\bibitem{stich2018local}
S.~U. Stich, ``Local sgd converges fast and communicates little,'' \emph{arXiv
  preprint arXiv:1805.09767}, 2018.

\bibitem{mcmahan}
H.~B. McMahan, E.~Moore, D.~Ramage, and B.~A. y~Arcas, ``Federated learning of
  deep networks using model averaging,'' \emph{ICLR Workshop Track}, 2016.

\bibitem{bonawitz2019towards}
K.~Bonawitz, H.~Eichner, W.~Grieskamp, D.~Huba, A.~Ingerman, V.~Ivanov,
  C.~Kiddon, J.~Konecny, S.~Mazzocchi, H.~B. McMahan \emph{et~al.}, ``Towards
  federated learning at scale: System design,'' \emph{arXiv preprint
  arXiv:1902.01046}, 2019.

\bibitem{yang}
\BIBentryALTinterwordspacing
Y.~Chen, X.~Sun, and Y.~Jin, ``Communication-efficient federated deep learning
  with asynchronous model update and temporally weighted aggregation,''
  \emph{arXiv preprint}, 2019. [Online]. Available:
  \url{https://arxiv.org/pdf/1903.07424.pdf}
\BIBentrySTDinterwordspacing

\bibitem{konevcny2016federated}
J.~Kone{\v{c}}n{\`y}, H.~B. McMahan, D.~Ramage, and P.~Richt{\'a}rik,
  ``Federated optimization: Distributed machine learning for on-device
  intelligence,'' \emph{arXiv preprint arXiv:1610.02527}, 2016.

\bibitem{lian2015asynchronous}
X.~Lian, Y.~Huang, Y.~Li, and J.~Liu, ``Asynchronous parallel stochastic
  gradient for nonconvex optimization,'' in \emph{Advances in Neural
  Information Processing Systems}, 2015, pp. 2737--2745.

\bibitem{zheng2017asynchronous}
S.~Zheng, Q.~Meng, T.~Wang, W.~Chen, N.~Yu, Z.-M. Ma, and T.-Y. Liu,
  ``Asynchronous stochastic gradient descent with delay compensation,'' in
  \emph{Proceedings of the 34th International Conference on Machine
  Learning-Volume 70}.\hskip 1em plus 0.5em minus 0.4em\relax JMLR. org, 2017,
  pp. 4120--4129.

\bibitem{Hogwild}
B.~Recht, C.~Re, S.~Wright, and F.~Niu, ``Hogwild: A lock-free approach to
  parallelizing stochastic gradient descent,'' in \emph{Advances in neural
  information processing systems}, 2011, pp. 693--701.

\bibitem{DeSaZhangOlukotunEtAl2015}
C.~M. De~Sa, C.~Zhang, K.~Olukotun, and C.~R{\'e}, ``{Taming the wild: A
  unified analysis of hogwild-style algorithms},'' in \emph{NIPS}, 2015, pp.
  2674--2682.

\bibitem{Leblond2018}
R.~Leblond, F.~Pedregosa, and S.~Lacoste-Julien, ``Improved asynchronous
  parallel optimization analysis for stochastic incremental methods,''
  \emph{JMLR}, vol.~19, no.~1, pp. 3140--3207, 2018.

\bibitem{nguyen2018sgd}
L.~M. Nguyen, P.~H. Nguyen, M.~van Dijk, P.~Richt{\'{a}}rik, K.~Scheinberg, and
  M.~Tak{\'{a}}c, ``{SGD} and hogwild! convergence without the bounded
  gradients assumption,'' in \emph{Proceedings of the 35th International
  Conference on Machine Learning, {ICML} 2018}, 2018, pp. 3747--3755.

\bibitem{Ghadimi2013Minibatch}
H.~Z. Saeed~Ghadimi, Guanghui~Lan, ``Mini-batch stochastic approximation
  methods for nonconvex stochastic composite optimization,'' \emph{arXiv
  preprint arxiv:1308.6594}, 2013.

\bibitem{van2020hogwild}
M.~van Dijk, N.~V. Nguyen, T.~N. Nguyen, L.~M. Nguyen, Q.~Tran-Dinh, and P.~H.
  Nguyen, ``Hogwild! over distributed local data sets with linearly increasing
  mini-batch sizes,'' \emph{arXiv preprint arXiv:2010.14763}, 2020.

\bibitem{george2019distributed}
J.~George and P.~Gurram, ``Distributed deep learning with event-triggered
  communication,'' \emph{arXiv preprint arXiv:1909.05020}, 2019.

\bibitem{hsieh2017gaia}
K.~Hsieh, A.~Harlap, N.~Vijaykumar, D.~Konomis, G.~R. Ganger, P.~B. Gibbons,
  and O.~Mutlu, ``Gaia: Geo-distributed machine learning approaching
  $\{$LAN$\}$ speeds,'' in \emph{14th $\{$USENIX$\}$ Symposium on Networked
  Systems Design and Implementation ($\{$NSDI$\}$ 17)}, 2017, pp. 629--647.

\bibitem{luping}
L.~Wang, W.~Wang, and B.~Li, ``Cmfl: Mitigating communication overhead for
  federated learning.'' \emph{IEEE International Conference on Distributed
  Computing Systems.}, 2019.

\bibitem{robbins1951stochastic}
H.~Robbins and S.~Monro, ``A stochastic approximation method,'' \emph{The
  annals of mathematical statistics}, pp. 400--407, 1951.

\bibitem{nguyen2018new}
L.~M. Nguyen, P.~H. Nguyen, P.~Richt{{\'a}}rik, K.~Scheinberg,
  M.~Tak{{\'a}}{\v{c}}, and M.~van Dijk, ``New convergence aspects of
  stochastic gradient algorithms,'' \emph{Journal of Machine Learning
  Research}, vol.~20, no. 176, pp. 1--49, 2019.

\end{thebibliography}



\end{document}